\documentclass[sigconf]{acmart}
\usepackage[utf8]{inputenc}
\usepackage{amsmath}
\usepackage{amsfonts}

\usepackage{amssymb}
\usepackage{geometry}
\usepackage{xcolor}
\usepackage{graphicx} 
\usepackage{subcaption}
\usepackage{subfiles}
\usepackage[ruled,vlined]{algorithm2e}
\usepackage{makecell}
\usepackage{enumitem}
\usepackage{colortbl} 

\title[Decision-focused Sensing and Forecasting]{Decision-focused Sensing and Forecasting for Adaptive and Rapid Flood Response: An Implicit Learning Approach}

\author{Qian Sun}
\email{qsun32@jhu.edu}
\affiliation{%
  \institution{Johns Hopkins University, USA \\
  HKUST, Hong Kong SAR, China}
  \country{}
}
\author{Graham Hults}
\email{ghults1@jhu.edu}
\affiliation{%
  \institution{Johns Hopkins University}
  \country{USA}
}
\author{Susu Xu}
\authornote{Corresponding author.}
\email{susuxu@jhu.edu}
\affiliation{%
  \institution{Johns Hopkins University}
  \country{USA}
}

\copyrightyear{2025}
\acmYear{2025}
\setcopyright{acmlicensed}
\acmConference[BUILDSYS '25]{The 12th ACM International Conference on
Systems for Energy-Efficient Buildings, Cities, and Transportation}{November
19--21, 2025}{Golden, CO, USA}
\acmBooktitle{The 12th ACM International Conference on Systems for
Energy-Efficient Buildings, Cities, and Transportation (BUILDSYS '25), November
19--21, 2025, Golden, CO, USA}
\acmDOI{10.1145/3736425.3770102}
\acmISBN{979-8-4007-1945-5/2025/11}
\begin{document}


\begin{abstract}
Timely and reliable decision-making is vital for flood emergency response, yet it remains severely hindered by limited and imprecise situational awareness due to various budget and data accessibility constraints. Traditional flood management systems often rely on in-situ sensors to calibrate remote sensing-based large-scale flood depth forecasting models, and further take flood depth estimates to optimize flood response decisions. However, these approaches often take fixed, decision task-agnostic strategies to decide where to put in-situ sensors (e.g., maximize overall information gain) and train flood forecasting models (e.g., minimize average forecasting errors), but overlook that systems with the same sensing gain and average forecasting errors may lead to distinct decisions. 

To address this, we introduce a novel decision-focused framework that strategically selects locations for in-situ sensor placement and optimize spatio-temporal flood forecasting models to optimize downstream flood response decision regrets. Our end-to-end pipeline integrates four components: a contextual scoring network, a differentiable sensor selection module under hard budget constraints, a spatio-temporal flood reconstruction and forecasting model, and a differentiable decision layer tailored to task-specific objectives. Central to our approach is the incorporation of Implicit Maximum Likelihood Estimation (I-MLE) to enable gradient-based learning over discrete sensor configurations, and probabilistic decision heads to enable differentiable approximation to various constrained disaster response tasks. Rather than minimizing proxy prediction loss, our method directly optimizes final task performance metrics such as evacuation and relocation cost. Extensive experiments on real-world flood scenarios demonstrate that our method substantially outperforms baselines in both state prediction accuracy and decision quality, offering a principled and effective solution for sensor-aware disaster management.

\end{abstract}
\maketitle

\section{Introduction}
Rapid and effective decision making in response to natural hazards (e.g. floods, hurricanes, wildfires, etc.), including massive evacuation, resource allocation, and facility relocation, is critical to reducing human and economic loss~\cite{xu2022seismic,wang2024scalable,zhou2018emergency}. To enable near-real-time decision-making, traditional paradigms often (1) first employ sensing systems to collect spatio-temporal data of disaster zones, (2) then use physical, statistical, or machine learning models to estimate or forecast spatio-temporal damage distribution, and finally (3) optimize various response decisions and dynamically update decisions as natural hazard damage evolves~\cite{wang2024near,chen2024soscheduler,li2023disasternet,huang2021systematic}. For example, in flood management, decision-makers often rely on multi-modal sensing data (e.g., precipitation, soil moisture, elevation, etc.) and machine learning models to estimate spatio-temporal distribution of flood depth, and then using these flood depth as decision parameters to optimize dynamic evacuation and facility relocation in hours to days before flooding to reduce loss. 

In particular, due to the large-scale and fast-changing nature of natural hazard damage, recent disaster sensing paradigms often involve multiple modalities ranging from remote sensing techniques and in-situ sensors deployed at key locations. Remote sensing data cover large-scale disaster zones, but often suffer from limited spatial resolution, infrequent updates, and poor signal quality~\cite{al2013major, amitrano2024flood}, making it difficult to accurately capture local flood dynamics. As complementary information sources, in situ sensors provide accurate and real-time point observations~\cite{munawar2022remote,tao2024review}, such as the stage and discharge of water, but often have limited spatial coverage due to high deployment and maintenance costs. In real-world practice, people often fuse large-scale, coarse-grained remote sensing data with sparse but accurate fine-grained in-situ measurements to jointly estimate and forecast the spatio-temporal distribution of flood depth using machine learning models~\cite{liu2025comprehensive, bukhari2024enhancing}, where in-situ point observations often play the role of calibrating remote sensing-based estimations. Therefore, the effectiveness of final response decisions are jointly determined by the quality of sensing measurements, forecasting models, and decision optimization algorithms.  To enable \emph{effective flood response systems}, people often need to \textbf{jointly} answer three key questions: \emph{(1) where to place in-situ sensors? (considering satellite imagery often has fixed characteristics) (2) how to adapt hazard/damage forecasting models with multi-modal sensing data? (3) how to optimize response decisions with forecasting results?}, by simultaneously considering their coupled impacts on final response decision quality.

However, existing flood management frameworks often isolate sensor deployment optimization and forecasting model optimization from actual disaster decision-making problems, resulting in sub-optimal outcomes. In fact, disaster responses involve a diverse spectrum of decision tasks, including assignment, matching, routing, and control, each needing tailored sensing strategies and forecasting models. 
Current approaches for sensing network optimization mostly prioritize maximizing \emph{average} information gain. Similarly, many data-driven forecasting models focus on minimizing \emph{average} forecasting errors on sampling datasets~\cite{li2023normalizing,
li2025rapid,li2025scalable}, but these objectives do not necessarily improve disaster response decisions~\cite{saunders2025data,gerolimos2025autonomous}. In real-world highly uncertain and resource-constrained scenarios, systems with the same sensing coverage and learning-based prediction accuracy may lead to distinct decisions and decision optimality. For example, achieving lower forecasting errors around critical infrastructures can significantly enhance resource allocation, even if overall average performance is unchanged. Traditional sensing objectives (e.g., aggregated information gain) and forecasting objectives (e.g., average training loss) may not coincide with optimal decision regrets, underscoring the need for a decision-focused, end-to-end framework to boost flood resilience.

To bridge this significant research gap, we introduce an end-to-end, implicit decision-driven framework that jointly learns to place sensors with limited budget and accessibility constraints, optimize flood estimation models, and make various flood response decisions to improve flood management effectiveness and efficiency. Specifically, our approach consists of four interconnected, decision problem-informed components: (1) a \emph{location scoring network} that assigns importance scores to candidate locations based on contextual features from satellite data; (2) a \emph{sensor sampling module} that samples a subset of locations for deployment under hard budget/accessibility constraints; (3) a \emph{spatio-temporal reconstruction and forecasting module} that reconstructs and predicts spatio-temporal flood depth across the entire watershed based on coarse-grained remote sensing measurements of environmental factors (gridded precipitation), sparse in-situ point observations (water depth and velocity), and geospatial features (digital elevation model, satellite-derived Manning's n surface roughness values, and satellite-derived distance to nearest waterbody); and (4) a \emph{differentiable decision head} that takes the predicted flood depth as input and outputs actionable plans tailored to specific flood response tasks. These four components are jointly trained and optimized to minimize the final decision regrets.



Key challenges in this decision-focused end-to-end learning framework arise from the inherently discrete and combinatorial nature of both sensor placement optimization and disaster response decision-making. Specifically, these tasks form combinatorial optimization problems characterized by cardinality constraints and non-differentiable discrete choices, thereby precluding direct application of gradient-based methods.  Secondly, optimal sensor subset selection from extensive candidate locations, as well as operational decisions like evacuation planning or strategic facility relocation, often involve exponentially large discrete search spaces due to the large-scale nature of natural hazard damage and decision-making. Moreover, joint optimization of these two problems with intermediate flood forecasting models further exacerbates computational complexity and intractability of feasible solutions. To overcome these challenges, our work has four major contributions:
\begin{itemize}[leftmargin=*]
\item We introduce a novel decision-focused, end-to-end framework that jointly optimizes sensing network deployment, damage estimation models, and disaster response decisions to enable effective, rapid, and scalable disaster management, using flood response as an example application scenario.
\item We design a novel location scoring network and sensor sampling module with \emph{Implicit Maximum Likelihood Estimation (I-MLE)}, to enable sensor placement optimization in a differentiable manner. With such design, the decision-aware gradients can propagate through the entire pipeline. 

\item To adapt to various disaster response tasks, we design and develop novel differentiable decision heads to address assignment and matching problems. In particular, we reformulate disaster response tasks into distributional optimization problems and introduce sinkhorn normalization and overflow penalty to respect constraints. Such decision heads enable differentiable information propagation from decision regrets to sensing and forecasting modules.

\item We evaluate our framework for a real-world event, the Offutt Air Force Base flood of March 2019, by leveraging remote sensing and in-situ monitoring data,  USGS geospatial data, and infrastructure system features. Through extensive experiments with two downstream tasks, we demonstrate that our framework learns to place sensors in valuable locations, forecasts flood depth in a spatio-temporal manner, and enables high-quality decisions under uncertainty. Our results suggest that our integrated, decision-focused sensing, forecasting, and decision-making framework substantially enhances the responsiveness and reliability of flood management systems.
\end{itemize}

\section{Related Works}
\subsection{Optimal Sensor Placement}
The optimal sensor placement problem has been explored in various application scenarios such as transportation systems\cite{gentili2018review,li2011reliable}, environmental monitoring\cite{lopez2023wireless,romano2021review}, etc. The task typically involves selecting a subset of candidate locations within a network or region to deploy sensors under a given budget or resource constraint, with the goal of maximizing information gain or improving state estimation accuracy. Specifically, one line of research adopts information-theoretic criteria such as entropy-based methods or mutual information based methods~\cite{papadimitriou2000entropy, said2000optimal} to select sensors that maximize relevance to the system state. Another line focuses on numerical approaches, including QR decomposition~\cite{kim2021greedy,yu2020optimal}, Fisher information matrix~\cite{kim2024effective,lee2021optimal}, and principal component analysis (PCA)~\cite{kerschen2004sensor}, which guide sensor selection based on algebraic structure.
\subsection{Decision Focused Learning}
Real-world decision-making problems under uncertainty, such as route planning and resource allocation, normally involves a predictive component which can be approached by a machine learning model to estimate the uncertain parameters, followed by a constrained optimization model which finds the optimal decision based on the predicted parameters. Such two-step formulation is defined as the predict-and-optimize paradigm.\cite{mandi2024decision,elmachtoub2022smart} However, the independent two-stage process often leads to suboptimal decisions even though the parameters lead to minimal prediction accuracy. Thus, decision-focused learning (DFL) is proposed to directly train the machine learning model for good decisions. The literature of DFL that addresses combinatorial optimization with constraint functions without specific form includes two prominent families. (i) Smoothing by Random Perturbations, which applies implicit regularization via noise and constructs smooth approximations of optimization mappings for backpropagation, e.g., perturb-and-MAP, DPO, and their variants \cite{berthet2020dpo,paulus2020sst,vlastelica2020blackbox,sahoo2023identity}. (ii) Differentiation of Surrogate Loss Functions, which proposes decision-aware surrogates (such as regret-aligned objectives) that reflect decision quality and yield gradients that are easy to compute for gradient-based training \cite{liu2021risk,mulamba2021contrastive,mandi2020spoilp}. Methods in the latter family construct regret-aligned surrogates \emph{only when the ground-truth cost vector} \(c\) \emph{is observed} (often together with an oracle or computable \(x^\star(c)\)); otherwise the regret is undefined and the surrogate cannot be instantiated \cite{mandi2024decision,liu2021risk}. 

\subsection{Spatio-Temporal Imputation and Forecasting}

Spatio-temporal models are widely adopted in both imputation and forecasting tasks, offering a unified framework to capture temporal dynamics and spatial dependencies. For imputation, recent deep learning models capture complex temporal-spatial correlations by learning such dependencies directly from data. Representative architectures include Convolutional, Recurrent, and Graph Neural Network hybrids, and Generative Adversarial Network-based imputers~\cite{asadi2019convolutional,yuan2022stgan,ding2025stp}, which can accommodate auxiliary inputs and irregular missing patterns. In terms of forecasting, recent models combine sequence encoders such as Diffusion models, and Transformers, with spatial modules like graph convolution or attention mechanisms.~\cite{zhang2024irregular,han2024bigst} These methods enable accurate multi-step predictions and support downstream applications such as traffic control, early warnings, and disaster response~\cite{sun2024spatio,sun2023hierarchical,sun2024crosslight}.

\section{Methodology}

\subsection{Problem Formulation}
We consider the problem of learning a \emph{discrete sensor sampling strategy} for in-situ flood monitoring, where the goal is to select a subset of sensor locations from a candidate set under a strict budget constraint. During a flood event, only a limited number of in-situ water monitoring sensors can be deployed, and each deployed sensor provides localized, high-fidelity observations that complement coarse-grained, remote-sensing data available across the entire region. The key challenge lies in determining which subset of candidate locations will yield the most useful information for the task, under uncertainty. Our objective is not merely to reconstruct the full flood state from partial observations, but to ensure that the resulting reconstruction supports effective \emph{downstream flood response decisions}. Therefore, we aim to learn a sensor sampling strategy that selects locations whose observations enable high-quality spatio-temporal inference and, more importantly, minimize the final task-specific loss incurred during decision-making. This requires modeling the interaction between sensor sampling, damage estimation, and decision outcomes in a principled, end-to-end manner. We formalize this end-to-end pipeline through three interconnected components, namely sensing, reconstruction and prediction, and decision-making.
\begin{figure}[htbp]
    \centering
    \includegraphics[width=\linewidth]{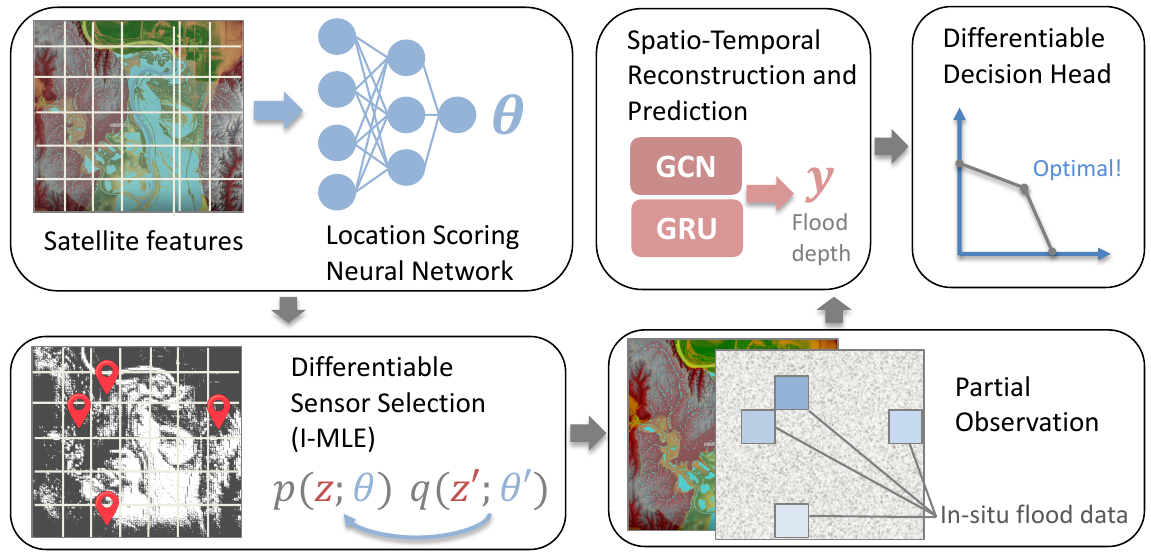}
    \caption{Framework Architecture.}
    \label{fig:framework}
\end{figure}



\noindent\textbf{\emph{Sensing. }}  Let $\mathcal{L} = \{l_1, \dots, l_N\}$ denote the set of candidate locations, where each location represents an unstructured cell in a feasible sampling region characterized by surface roughness values and hydrodynamic properties. These locations serve as the spatial units over which sensor placement is performed. At each location $l_i \in \mathcal{L}$, we have access to remote-sensing based features $x_i \in \mathbb{R}^d$ (e.g., precipitation, elevation, distance to nearest water body), which are continuously available but lack the precision required for accurate flood state estimation, motivating the need for in-situ sensor deployment to collect water height and velocity data. We denote the binary sensor placement vector by $\mathbf{z} = (z_1, \dots, z_N) \in \{0,1\}^N$, where $z_i = 1$ indicates that a sensor is deployed at location $i$, and $z_i = 0$ otherwise. The deployment is constrained by a fixed budget $K \ll N$, i.e., $\sum_{i=1}^{N} z_i = K.$ Hence, we can formulate the sensor placement task as a combinatorial optimization problem, where the goal is to select a subset of $K$ locations from $N$ candidates to maximize the informativeness of partial observations for downstream decision-making. Let $\theta \in \mathbb{R}^N$ denote a real-valued score vector produced by a sensor scoring network, reflecting the estimated utility of deploying a sensor at each location. The optimal sensor configuration $\mathbf{z}^*$ is then defined as the solution to the following constrained maximization:
\begin{equation}
\label{eq:combinatorial_argmax}
\mathbf{z}^* = \arg\max_{\mathbf{z} \in \mathcal{C}_K} \langle \theta, \mathbf{z} \rangle, \; \mathcal{C}_K = \left\{ \mathbf{z} \in \{0,1\}^N \mid \sum_{i=1}^N z_i = K \right\}.
\end{equation}

This objective selects the top-$K$ entries of $\theta$, corresponding to the most informative sensor locations. However, the argmax operation over a combinatorial space is non-differentiable, making it incompatible with standard backpropagation. To address this, we recast the problem as sampling from a discrete exponential family distribution over $\mathcal{C}_K$, parameterized by $\theta$, and employ perturbation-based sampling methods to enable gradient flow through the discrete layer. This transformation allows the sensor placement problem to be seamlessly integrated into an end-to-end differentiable learning pipeline.

\noindent\textbf{\emph{Reconstruction and Prediction. }}
Given a sensor configuration $\mathbf{z}^*$, let $\mathbf{h}$ denote approximate in-situ sensor observations using hydraulic model outputs for depth and velocity, we obtain partial observations $\mathbf{o}=[\,\mathbf{x};\, \mathbf{z}^*\!\odot\!\mathbf{h}\,]$, i.e., the in-situ term is masked when $z_i^*=0$. Since the observed information $\mathbf{o}$ is sparse, reconstructing the full environmental state is essential for effective global decision-making. We propose to feed these sparse and partially observed signals into a spatio-temporal reconstruction and forecasting module, which computes a dense estimate $\hat{\mathbf{y}} \in \mathbb{R}^N$ of the underlying environmental state, such as full-region flood depth.


\noindent\textbf{\emph{Decision Making. }}
The reconstructed environmental state $\hat{\mathbf{y}}$ serves as the input for downstream decision-making tasks. In particular, we focus on two representative tasks including evacuation planning and aircraft relocation. 

The evacuation planning task involves allocating evacuees to a set of precomputed evacuation routes to designated shelters with capacity constraints, which can be formally formulated as a \textit{capacitated assignment problem}. We construct a directed evacuation graph $G = (V, E)$, where each edge $e \in E$ represents a traversable junction, and its travel cost $c_e$ is a monotonic function of flood depth and edge length. Let $\mathcal{P}$ denote the set of precomputed feasible evacuation paths from the origin to fixed destination nodes. For each path $p \in \mathcal{P}$, let $d_p \in \mathbb{Z}_{\geq 0}$ denote the number of evacuees to be assigned to path $p$, and let $c_p = \sum_{e \in p} c_e$ be the total travel cost of $p$, the evacuation problem is then formulated as the following integer linear program:
\begin{align}
\min & \sum_{p \in \mathcal{P}} d_p \cdot c_p, \quad d_p \in \mathbb{Z}_{\geq 0}, \\
\text{s.t.} \quad
& \sum_{p \in \mathcal{P}} d_p = D, \quad \sum_{p \in \mathcal{P}_j} d_p \leq C_j, \quad \forall j \in \mathcal{S}.
\end{align}
Here, \( D \) is the total evacuation demand from the origin node, and \( C_j \) denotes the capacity of destination shelter \( j \in \mathcal{S} \). Each path \( p \in \mathcal{P} \) terminates at some shelter, and \( \mathcal{P}_j \subseteq \mathcal{P} \) denotes the set of paths leading to shelter \( j \). The objective minimizes the total evacuee-weighted travel cost, while ensuring that all demand is allocated and no shelter exceeds its capacity.

In the aircraft relocation task, our goal is to assign each aircraft stationed at a flood-vulnerable facility to a designated safe hangar within the region, while minimizing the total relocation cost and ensuring one-to-one matching feasibility. This can naturally be formulated as a \textit{bipartite matching problem}. Specifically, let $\mathcal{A} = \{A_1, \dots, A_M\}$ denote the set of $M$ aircraft, and $\mathcal{H} = \{H_1, \dots, H_L\}$ denote the set of L available hangars. We define a bipartite graph between $\mathcal{A}$ and $\mathcal{H}$, where each edge $(i, j)$ carries a flood aware relocation cost $c_{ij}$ from aircraft $A_i$ to hangar $H_j$. We want to assign binary decision variables $d_{ij} \in \{0,1\}$, where $d_{ij} = 1$ indicates that aircraft $A_i$ is assigned to hangar $H_j$. The optimization problem is then defined as:
\begin{equation}
    \min \sum_{i=1}^{M} \sum_{j=1}^{L} c_{ij} \cdot d_{ij}, \quad d_{ij} \in \{0,1\},
\end{equation}
\begin{equation}
    \text{s.t.} \sum_{j=1}^{L} d_{ij} \leq 1, \forall i \in [M], \quad 
  \sum_{i=1}^{M} d_{ij} \leq 1, \forall j \in [L].
\end{equation}

This formulation encodes a one-to-one matching between aircraft and hangars. The objective minimizes total relocation cost while ensuring each aircraft is assigned to at most one hangar and no hangar receives more than one aircraft.
 
In both tasks, decisions are modeled using binary or integer-valued variables. This discrete and discontinuous nature of the solution spaces makes direct integration into gradient-based learning pipelines highly challenging, preventing end-to-end training of predictive models when supervision is provided solely through final decision quality. To address this, we propose task-specific loss functions and design differentiable decision network heads that enable the learning pipeline to align predictions with decision performance. Further methodological details are provided in the following section.

\subsection{Model Architecture}
Given the problem setup described above, we explain details of our framework design in this section. 

\subsubsection{Location Scoring Network}

The location scoring network serves as the first component of our pipeline, responsible for assigning an importance score to each candidate location based on available remote-sensing inputs. Formally, given input features $\mathbf{x}$, the scoring network computes a score vector $\theta = (\theta_1, \dots, \theta_N) \in \mathbb{R}^N$ through a learnable function $h_v : \mathbb{R}^{N \times d_h} \rightarrow \mathbb{R}^N$, parameterized as a neural network.

The resulting score vector $\mathbf{\theta}$ parameterizes the distribution from which sensor configurations $\mathbf{z}$ are sampled in the next stage. Higher values of $\theta_i$ indicate a greater estimated utility of deploying a sensor at location $i$ for downstream decision quality. Importantly, this scoring function is trained end-to-end using gradients from the final task loss, ensuring that sensor utility is evaluated not by reconstruction quality alone, but by ultimate decision impact.

\subsubsection{Sensor Sampling Module}
In order to learn a task-aware sampling strategy over combinatorial sensor configurations, we require a principled way to define and optimize a distribution over all valid sensor placements. Therefore, we model the selection process as sampling the binary placement vector $\mathbf{z}$ from a parameterized distribution, where the constraint ensures that exactly $K$ sensors are selected. This probabilistic formulation enables us to reason about uncertainty and incorporate stochasticity into the learning process. To this end, we represent the sensor sampling distribution using a \emph{discrete exponential family}~\cite{niepert2021implicit}, defined as $p(\mathbf{z}; \theta) \propto \exp(\langle \mathbf{z}, \theta \rangle)$, 
\text{subject to Equation \ref{eq:combinatorial_argmax}}, where $\mathbf{\theta}$ is obtained from the location scoring network. The inner product $\langle \mathbf{z}, \theta \rangle$ reflects the total score of a configuration, i.e., placements that activate high-scoring locations under $\theta$ are more likely to be sampled.

The exponential family form provides a principled probabilistic model over combinatorial structures like top-$K$ subsets in our problem setup, without requiring continuous relaxations~\cite{jang2016categorical}. Moreover, it enables efficient sampling via \emph{perturb-and-MAP} methods~\cite{papandreou2011perturb,kool2019stochastic}. Particularly, instead of sampling directly from the distribution, we add structured random noise $\epsilon \sim \rho$ to the score vector and perform a top-$K$ argmax selection, where the noise $\epsilon$ is drawn from a special distribution such as \emph{Sum-of-Gamma} \cite{niepert2021implicit}, designed to approximate Gumbel noise under subset constraints: $\mathbf{z^*} = \arg\max_{\mathbf{z} \in \mathcal{C}_K} \langle \mathbf{z}, \theta + \epsilon \rangle.$ This perturb-and-MAP scheme makes the discrete sampling operation differentiable in expectation, and thus allows gradient-based learning of the scoring function. To enable end-to-end training with respect to downstream decision performance, we adopt the \emph{Implicit Maximum Likelihood Estimation (I-MLE)} technique~\cite{niepert2021implicit}. The core idea is to define a surrogate \emph{target distribution} $q(\mathbf{z}; \theta')$ that assigns higher probability to configurations yielding better task outcomes, and optimize the model distribution $p(\mathbf{z}; \theta)$ to match it. Specifically, we define:
\begin{equation}
\theta' = \theta - \lambda \nabla_{\mathbf{z}} \mathcal{L}_{\text{task}}\left( f_\phi\left( g_\psi\left( \mathbf{o}(\mathbf{z}) \right) \right), \mathbf{y} \right),
\end{equation}
where the gradient is taken with respect to the sampled configuration $\mathbf{z}$, while $\lambda > 0$ is a scalar hyperparameter that controls the magnitude of the task-aligned gradient step used to construct the target distribution, and \(g_\psi(\cdot)\) and \(f_\phi(\cdot)\) stand for the spatio-temporal network and differentiable decision head. I-MLE then approximates the gradient of the KL divergence between $p$ and $q$ via the difference of two MAP samples:
\begin{equation}
\nabla_\theta \mathcal{L} \approx \text{MAP}(\theta + \epsilon) - \text{MAP}(\theta' + \epsilon),
\end{equation}
where both terms are differentiable via backpropagation through the scoring network. This allows the sensor placement strategy to be directly trained based on its effect on the final decision loss, rather than intermediate surrogates.

The I-MLE framework can be perceived as implicitly maximizing the likelihood of high-quality sensor placements under the model distribution $p(\mathbf{z}; \theta)$. Specifically, I-MLE minimizes the KL divergence between a loss-improving target distribution $q(\mathbf{z}; \theta')$ and the current model distribution:
\begin{equation}
\label{eq:kl-objective}
\mathrm{KL}(q \| p) = \mathbb{E}_{\mathbf{z} \sim q}[\log q(\mathbf{z; \theta'})] - \mathbb{E}_{\mathbf{z} \sim q}[\log p(\mathbf{z}; \theta)].
\end{equation}
Since the entropy term $\mathbb{E}_{\mathbf{z} \sim q}[\log q(\mathbf{z}; \theta)]$ is independent of $\theta$, minimizing the KL divergence reduces to maximizing the expected log-likelihood of configurations sampled from the target distribution: $\max_{\theta} \; \mathbb{E}_{\mathbf{z} \sim q(\theta')} [\log p(\mathbf{z}; \theta)].$ Instead of maximizing likelihood on ground-truth samples, I-MLE shifts the target distribution $q$ toward configurations that yield lower task loss, and encourages the model distribution $p$ to align with it. Through this implicit likelihood matching, the model learns to prioritize sensor configurations that are high-rewarding, while avoiding the need to compute expectations or gradients through the task loss under $p$.

\subsubsection{Spatio-temporal Reconstruction and Forecasting}
To leverage both spatial correlations across cells within the region and temporal evolution across multiple timesteps for flood depth reconstruction and prediction, we adopt a dual spatial and temporal architecture. Specifically, at each time step $t = 1, \dots, T$, we first construct a feature vector at each location $i$ as: $\tilde{x}_{t}^{(i)} = \text{Concat}(o_t^{(i)}, z_t^{(i)}) \in \mathbb{R}^{d_o+1}$, which is first passed through a single-layer Graph Convolutional Network (GCN) to aggregate neighborhood information across the spatial graph $\mathcal{G} = (\mathcal{V}, \mathcal{E})$:
$S_t = \sigma\left( A \tilde{X}_t W_\text{S} \right),$, where $S_t \in \mathbb{R}^{N \times h}$, and $\mathcal{G} = (\mathcal{V}, \mathcal{E})$ denotes the spatial graph over the study region, with each node $v_i \in \mathcal{V}$ corresponding to a cell and each edge $(v_i, v_j) \in \mathcal{E}$ representing spatial or hydrological adjacency (e.g., upstream-downstream flow or proximity) among the cells. $A \in \mathbb{R}^{N \times N}$ is the normalized adjacency matrix derived from $\mathcal{G}$, encoding the structural connectivity among nodes. $W_\text{S} \in \mathbb{R}^{(d_o+1) \times h}$ is a trainable weight matrix, and $\sigma(\cdot)$ is a non-linear activation function such as ReLU.

In order to further capture temporal dynamics of water height observations at each location across timesteps, we then feed the spatially encoded features $S_t$ into a Gated Recurrent Unit (GRU), which operates independently at each cell location $v_i$ and regressively updates its hidden state $h_t^{(i)} \in \mathbb{R}^{h}$ via: $h_t^{(i)} = \text{GRU}\left(S_t^{(i)}, h_{t-1}^{(i)}\right)$, for $t \in \{1, 2, ..., T\}$, $i \in \{1, 2, ..., N\}$. After the final timestep $T$, we obtain the hidden state matrix $H \in \mathbb{R}^{N \times h}$, which is passed through a fully connected decoder to produce flood depth estimates via $\hat{y} = H W_\text{MLP} + b_\text{MLP}$.

\subsubsection{Differentiable Decision Head}
To enable end-to-end optimization toward task-level objectives, we design differentiable decision heads tailored to the structure of each downstream problem. These heads take $\hat{y}$ as input, and output continuous approximations of discrete decision variables, enabling gradient flow back to aforementioned model components.

\textit{Capacitated Assignment Head.} For the evacuation planning task, a fixed number of evacuees must be allocated to a set of precomputed evacuation routes $P$ under capacity constraints. Given $\hat{y}$, we apply mean pooling across spatial locations to obtain a global descriptor: $\bar{y} = \frac{1}{N} \sum_{i=1}^N \hat{y}_i$, which is then passed through a feedforward network to produce route scores $s \in \mathbb{R}^P$. A softmax transformation then yields the predicted allocation proportions following:
$p = \mathrm{softmax}(s) \in [0,1]^P.$
Multiplying by the total number of evacuees $D$, we can obtain the predicted assignment $d = D \cdot p$, where $\sum_{i=1}^M d_i = D$. To supervise this soft assignment, we define a decision-focused loss that penalizes both evacuation cost and capacity violations:
\begin{equation}
\mathcal{L}_{\text{assign}} = \sum_{i=1}^P c_i \cdot d_i + \gamma_{\text{assign}} \cdot \sum_{j=1}^S \mathrm{ReLU}\left( \sum_{i \in \mathcal{P}_j} d_i - u_j \right),
\label{eq:assign_loss}
\end{equation}
where \( c_i \) denotes the estimated traversal cost of route \( i \) (computed based on groundtruth flood levels), \( d_i \) is the number of evacuees assigned to route \( i \), \( u_j \) is the capacity of shelter \( j \), and \( \mathcal{P}_j\) denotes the set of routes that lead to shelter \( j \). The scalar \( \gamma_{\text{assign}} \) controls the penalty for capacity overflow. This formulation yields a differentiable surrogate for the original discrete assignment problem, enabling gradients from decision quality to flow back to upstream modules.

\textit{Bipartite Matching Head.}  
For aircraft relocation, we formulate the task as a bipartite matching between $M$ aircraft and $L$ candidate hangars, subject to one-to-one constraint. To enable differentiable learning, we need to obtain a soft assignment matrix $\mathbf{P} \in [0,1]^{M \times L}$ that approximates a binary matching.

Specifically, we leverage MLP to produce a score matrix $\mathbf{S} \in \mathbb{R}^{M \times L}$. In order to obtain a differentiable matching, we apply \emph{Sinkhorn normalization}~\cite{altschuler2017near} to the negative score matrix $\log \alpha := -\mathbf{S}$, projecting it onto the Birkhoff polytope. The iterative procedure alternates between row and column normalization in the log domain:
\begin{align}
\log \boldsymbol{\alpha}^{(t+1)} &= \log \boldsymbol{\alpha}^{(t)} 
- \log \sum_{j=1}^L \exp(\log \boldsymbol{\alpha}^{(t)}_{ij}), \\
\log \boldsymbol{\alpha}^{(t+2)} &= \log \boldsymbol{\alpha}^{(t+1)} 
- \log \sum_{i=1}^M \exp(\log \boldsymbol{\alpha}^{(t+1)}_{ij}).
\end{align}
After $\tau$ iterations, we obtain the final soft assignment:
\begin{equation}
\mathbf{P} = \exp(\log \boldsymbol{\alpha}^{(\tau)}), \quad 
\sum_{j=1}^L P_{ij} \approx 1, \quad \sum_{i=1}^M P_{ij} \approx 1.
\end{equation}

The resulting matrix $\mathbf{P}$ enables a smooth approximation of discrete matchings, and is supervised via a task-specific loss:
\begin{align}
\mathcal{L}_{\text{match}} =\; & \sum_{i,j} c_{ij} P_{ij} \nonumber + \gamma_{match} \cdot 
\sum_{j=1}^L \mathrm{ReLU}\left( \sum_{i=1}^M P_{ij} - u_j \right) \\
& + \gamma_{match} \cdot \sum_{i=1}^M \mathrm{ReLU}\left( \sum_{j=1}^L P_{ij} - 1 \right).
\label{eq:match_loss}
\end{align}
where $c_{ij}$ is the flood-aware relocation cost and $u_j$ is the capacity of hangar $j$, and $\gamma_{match}$ controls the weight of the contribution of the constraint violation loss. This formulation allows end-to-end training by aligning upstream predictions with structured downstream decisions.



\subsection{Training Pipeline}

We train the full model end-to-end to optimize evacuation performance under partial observability. We explain the training pipeline in terms of two processes: a forward pass that simulates the sensing, forecasting, and decision-making process, and a backward pass that applies decision-focused I-MLE-based surrogate gradients to update the sensor placement policy based on final task outcomes.

\subsubsection{Forward Pass}

At each training iteration, the forward passing consists of the following steps:

\begin{itemize}[leftmargin=*]
    \item \textbf{Location Scoring}: The location scoring network computes the score vector $\theta_t = h_v(x_t)$ for candidate sensor locations.

    \item \textbf{Sensor Selection}: Using perturb-and-MAP sampling, the agent selects $k$ sensor locations, following Equation \ref{eq:combinatorial_argmax}.

    \item \textbf{Partial Observation}: Partial hydraulic depth readings are obtained $o_t = \Omega(z_t)$ at the selected locations.

    \item \textbf{Full State Reconstruction and Prediction}: Leveraging the spatio-temporal module, a full state of flood depth is reconstructed, i.e., $\hat{y}_t=g_\psi({o_t})$.

    \item \textbf{Decision Making}: Given the predicted flood depth $\hat{y}_t$, the decision head is applied to produce a differentiable distribution over discrete decision variables $a_t$.

    \item \textbf{Evaluation}: The resulting soft decision vector $a_t$ is used to evaluate task loss $\mathcal{L}_{\text{task}}(a_t, y_t)$, which is differentiable with respect to upstream neural networks.

\end{itemize}

\subsubsection{Backpropagation}
At each training step, the following process is used to estimate a task-aware gradient for the sensor scoring network:
\begin{itemize}[leftmargin=*]
    \item \textbf{Constructing a Loss-Informed Target Distribution}: After the forward pass, we first compute the downstream task loss $\mathcal{L}_{\text{task}}(a_t, y_t)$ based on the differentiable downstream decision vector $a_t$ and the groundtruth flood state $y_t$. To guide sensor placement toward decisions that improve downstream outcomes, we construct a \textit{loss-informed score vector} $\theta'_t$ that reflects how the sensor selection should change to reduce the task loss. Specifically, we perform a first-order adjustment of the original sensor logits $\theta_t$:
    \begin{equation}
    \theta'_t = \theta_t - \lambda \cdot \nabla_{\mathbf{z}_t} \mathcal{L}_{\text{task}}(a_t, y_t)
    \end{equation}
    where $\lambda > 0$ is a step size, and the gradient is computed via automatic differentiation through the decision modules. This adjustment approximates the gradient descent direction in sensor space that most reduces task cost, without requiring direct backpropagation through discrete sampling.

    \item \textbf{Target MAP Sampling}:  
    Using the same noise realization as in the forward pass, we perform a second perturb-and-MAP sampling step to obtain a new discrete configuration:
    \begin{equation}
    z'_t = \arg\max_{z \in \mathcal{C}_k} \langle z, \theta'_t + \epsilon_t \rangle.
    \end{equation}
    This yields a task-improving structure drawn from a distribution that favors lower task loss, but remains differentiable via shared noise.

    \item \textbf{I-MLE Gradient Estimation}:  
    The gradient of the KL divergence between the model distribution $p(\cdot; \theta_t)$ and the target distribution $q(\cdot; \theta'_t)$ is estimated by the difference of the two MAP samples:
    \begin{equation}
    \hat{\nabla}_{\theta_t} \mathcal{L}_{\text{task}} = z_t - z'_t.
    \end{equation}
    This difference indicates the direction in which the score vector $\theta_t$ should be updated to match the better-performing target distribution.

    \item \textbf{Backpropagation to Scoring Network}:  
    Finally, the estimated gradient is propagated through the sensor scoring network using the chain rule:
    \begin{equation}
    \hat{\nabla}_{v} = \frac{\partial \theta_t}{\partial v} \cdot \hat{\nabla}_{\theta_t}.
    \end{equation}
    This step successfully updates the parameters $v$ of the scoring network $h_v$ to improve future sensor selection decisions in accordance with task-level outcomes.
\end{itemize}

This I-MLE-based training scheme enables the model to reason about the long-term consequences of its discrete placement decisions, while sidestepping the need for differentiable relaxations. Crucially, it allows credit assignment to be made through structured actions that influence information acquisition and decision quality in a nontrivial way. We provide the pseudocode in Algorithm \ref{alg1}. 

\begin{algorithm}[!h]
\caption{End-to-End Training Framework}
\label{alg:imle_dfl_training}
\KwIn{Satellite features $x_t$, groundtruth flood map $y_t$, sensor budget $K$.}
\KwOut{Gradient of sensor scoring network parameters $\hat{\nabla}_v$.}

\For{each training iteration}{
    $\theta_t \gets h_v(x_t)$ \tcp*{Compute location scores}
    $\boldsymbol{\epsilon}_t \sim \text{Gumbel}(0,1)^N$ \tcp*{Perturb}
    $\mathbf{z}_t \gets \arg\max_{\mathbf{z} \in \mathcal{C}_k} \langle \mathbf{z}, \theta_t + \boldsymbol{\epsilon}_t \rangle$ \tcp*{MAP}
    $o_t \gets \Omega(\mathbf{z}_t)$ \tcp*{Partial observation}
    $\hat{y}_t \gets g_\psi(o_t)$ \tcp*{Flood estimate}
    $\mathbf{a}_t \gets \text{DecisionHead}(\hat{y}_t)$ \tcp*{Decision making}
    $\mathcal{L}_{\text{task}}$ following Equations \ref{eq:assign_loss} and  \ref{eq:match_loss}\tcp*{Task loss}

    $\nabla_{\mathbf{z}_t} \mathcal{L}_{\text{task}} \gets \text{Autodiff}(\mathcal{L}_{\text{task}})$ \tcp*{Gradient estimation}
    $\theta'_t \gets \theta_t - \lambda \cdot \nabla_{\mathbf{z}_t} \mathcal{L}_{\text{task}}$ \tcp*{Loss-informed score}
    $\mathbf{z}'_t \gets \arg\max_{\mathbf{z} \in \mathcal{C}_k} \langle \mathbf{z}, \theta'_t + \boldsymbol{\epsilon}_t \rangle$ \tcp*{Target sample}
    $\hat{\nabla}_{\theta_t} \gets \mathbf{z}_t - \mathbf{z}'_t$ \tcp*{I-MLE surrogate gradient}
    $\hat{\nabla}_v \gets \frac{\partial \theta_t}{\partial v} \cdot \hat{\nabla}_{\theta_t}$ \tcp*{Gradient descent}
}
\label{alg1}
\end{algorithm}

\section{Experiments}
\subsection{Flood Simulation Region of Interest}
\begin{figure}[htbp]
    \centering
    \begin{subfigure}[t]{0.49\linewidth}
        \centering
        \includegraphics[width=\linewidth]{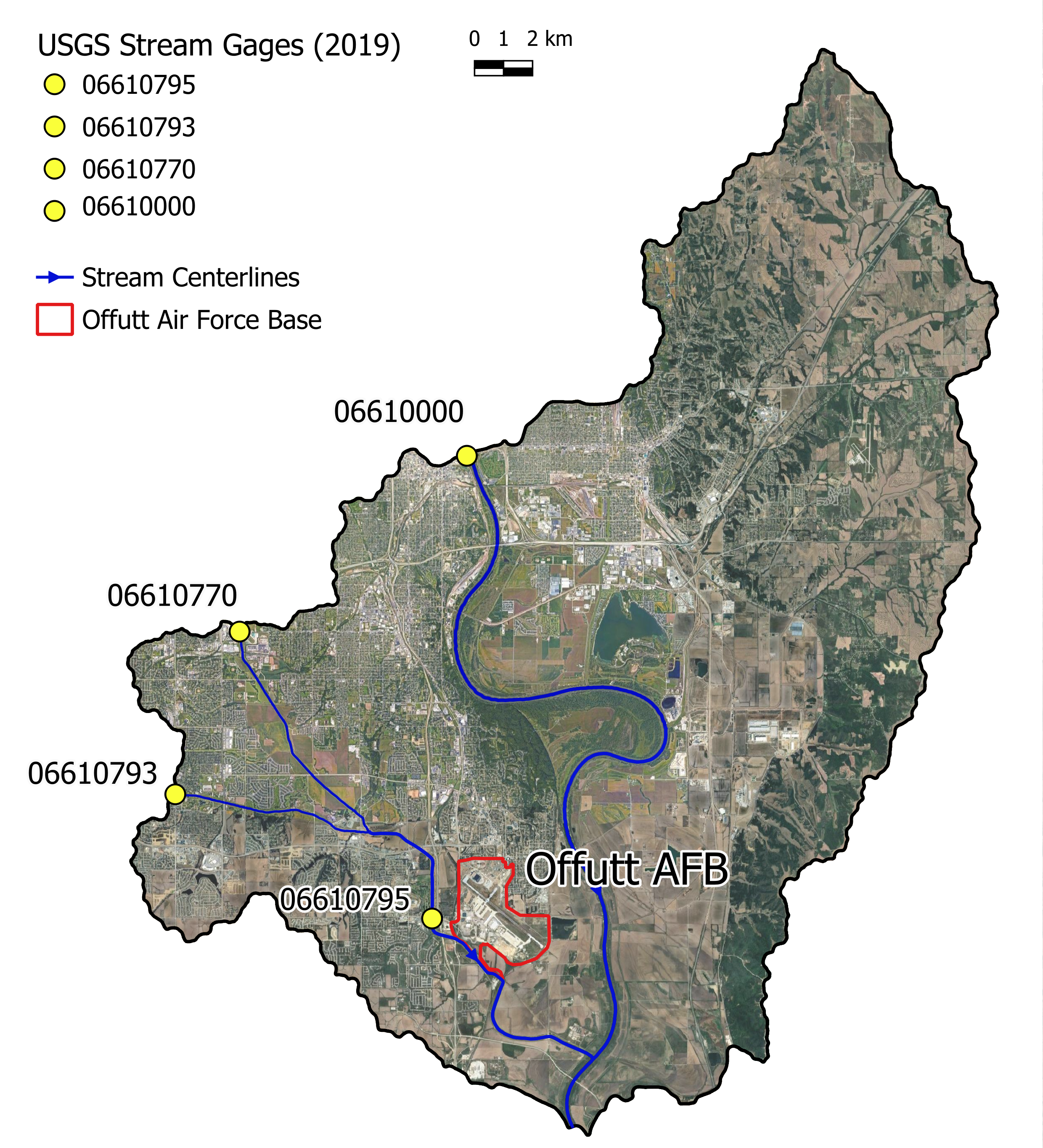}
        \caption{Watershed for evaluating the sensor placement problem.}
        \label{fig:watershed}
    \end{subfigure}
    \hfill
    \begin{subfigure}[t]{0.49\linewidth}
        \centering
        \includegraphics[width=\linewidth]{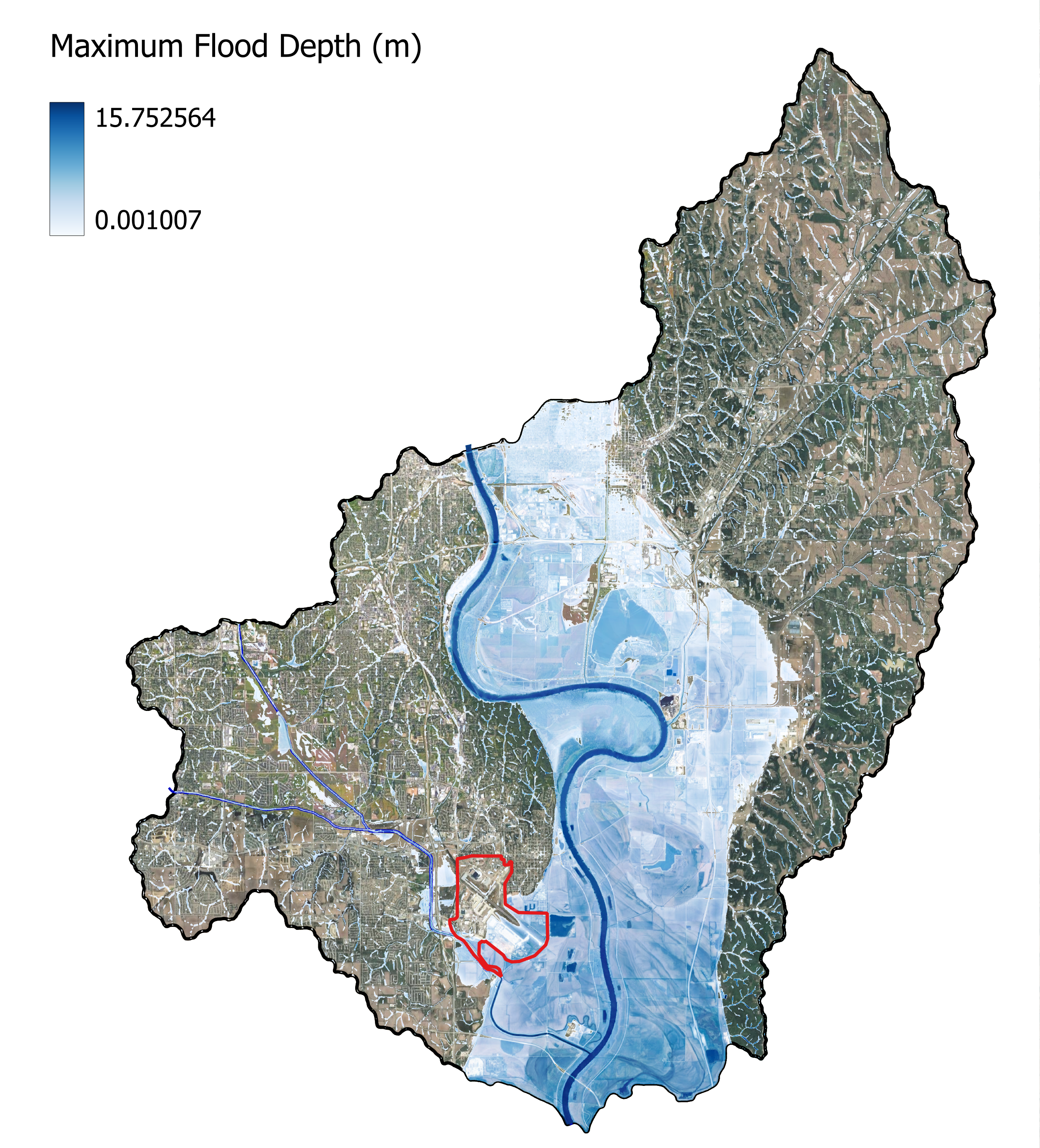}
        \caption{Maximum flood depth on March 19, 2019.}
        \label{fig:max_flood_depth}
    \end{subfigure}
    \caption{Study region and hydrological conditions.}
    \label{fig:study_region_and_flood}
\end{figure}

In March 2019, a combination of record snowfall, an irregular temperature spike, and 2-3 inches of rainfall caused a 1000-year flood event at Offutt Air Force Base, located south of Omaha, Nebraska \citep{hobza2023floodwater}. The Missouri River overtopped its banks and levee adjacent to the base, flooding roughly 130 military facilities, and causing \$700 million in damages, \$1.6 billion in reconstruction costs, and the relocation of flight operations to Lincoln, Nebraska for 18 months \citep{offutt_flood_rebuild_2024, lincoln_airport_growth_2024}.

We investigate the utility of our decision-focused sensor placement by simulating this flood event. To successfully capture hydrodynamics and decision-making objectives, we delineate the area illustrated in Figure \ref{fig:watershed} which includes observed discharge and water height data at inlet locations along Big Papillion Creek, Papillion Creek, and the Missouri River.


In all downstream tasks, sensors must be placed within a region feasible for measuring discharge or water height. We extract locations from the US Geological Survey National Hydrography Dataset \footnote{\url{https://www.sciencebase.gov/catalog/item/5a58a5e3e4b00b291cd69465}} water bodies feature layer to define feasible sensor placement space. Downstream resources and operational constraints, such as evacuated residents, critical infrastructure, aircraft and sandbag stocks, are concentrated within the Offutt Air Force Base (AFB)\footnote{{\url{https://www.offutt.af.mil/}}} region, spanning 11.3 square kilometers. Our aim is to leverage a limited number of strategically placed sensors over the watershed in Figure \ref{fig:watershed} to infer the full flood state, based on which we make optimized decisions for emergency response within the airbase region.

\subsection{Dataset Description}

We utilize the US Army Corps of Engineers 2D HEC-RAS software to simulate the 2019 flood event and develop a ground-truth dataset for flood depth reconstruction. HEC-RAS is the principal model used by FEMA’s National Flood Insurance Program and NOAA’s Advanced Hydrologic Prediction Service \citep{afshari2018comparison}. Geo-spatial inputs for the HEC-RAS 2D model include terrain data, a computational mesh, boundary condition lines, initial flow conditions, and meteorological data.

\begin{itemize}[leftmargin=*]
    \item \textit{Terrain.} We develop a Digital Elevation Model (DEM) mosaic by fusion of 1-meter LiDAR GeoTIFF of the Air Force base with a 10-meter GeoTIFF of the remaining area. The 1-meter LiDAR file was provided via email communications with the USGS and the 10-meter DEM was acquired through the USGS National Map Downloader \footnote{{\url{https://apps.nationalmap.gov/downloader/}}}.

    \item \textit{Mesh.} We generate a heterogeneous computational mesh file containing 131,190 cells. Stream layers from the NHD Plus dataset were utilized as break lines to refine mesh structure and spatially varied Manning’s N surface roughness values were applied to regulate flow across all cell faces. Manning's N values were determined through a reclassification of the 2019 National Land Cover Database \citep{usgs_nlcd}.

    \item \textit{Control Specifications.} We simulate the flood event at hourly intervals from 01 Mar 2019 00:00 through 31 Mar 2019 24:00.

    \item \textit{Boundary Conditions.} Time-series discharge data from USGS stream gage stations 06610000, 06610770, 06610793, and 06610795, sourced from the U.S. Geological Survey National Water Information System~\citep{usgs_nwis}, are utilized to model historic flow conditions at and through the boundary condition points into the meshed region.

    \item \textit{Initial Conditions.} We apply a model warm-up time of 4 hours with initial surface water elevation set to 289 meters based on local elevation values. This allows water to populate channels and water body surfaces prior to running the simulation and prevents a dry-start. 

    \item \textit{Meteorological Data.} We utilize gridded incremental precipitation data at 800-meter, 1-hour resolution from NOAA’s Analysis of Record for Calibration (AORC). \footnote{ \url{https://registry.opendata.aws/noaa-nws-aorc/}.}

\end{itemize}

The USGS previously conducted a steady-flow simulation for this area in 2023 of Papillion Creek \citep{hobza2023floodwater}. We newly model an unsteady simulation of both Papillion Creek and the Missouri River floodplain utilizing the 2D Diffusion Wave Equations. These equations quickly approximate the full Shallow Water Equations, ensuring conservation of mass and momentum, simulating hydrograph routing, wave propagation, and storage effects \citep{hecras2021}.
Figure~\ref{fig:max_flood_depth} represents peak flood inundation for this area occurring at 06:00 March 19, 2019.

\subsection{Flood Response Tasks}
\begin{figure}[htbp]
    \centering
    \begin{subfigure}[t]{0.49\linewidth}
        \centering
        \includegraphics[width=\linewidth]{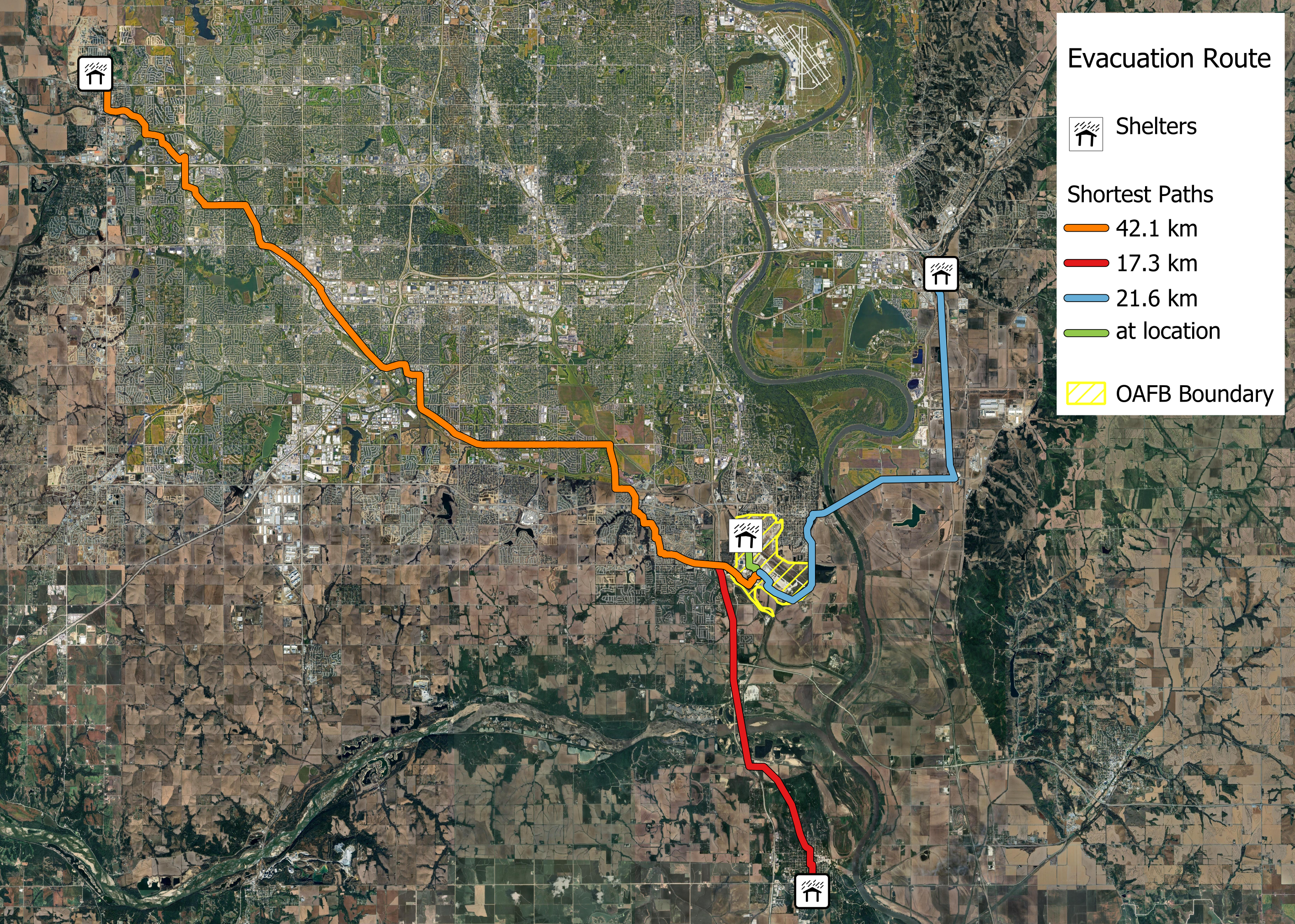}
        \caption{Candidate shelters and example candidate routes for the evacuation task.}
        \label{fig:evacuation}
    \end{subfigure}
    \hfill
    \begin{subfigure}[t]{0.49\linewidth}
        \centering
        \includegraphics[width=\linewidth]{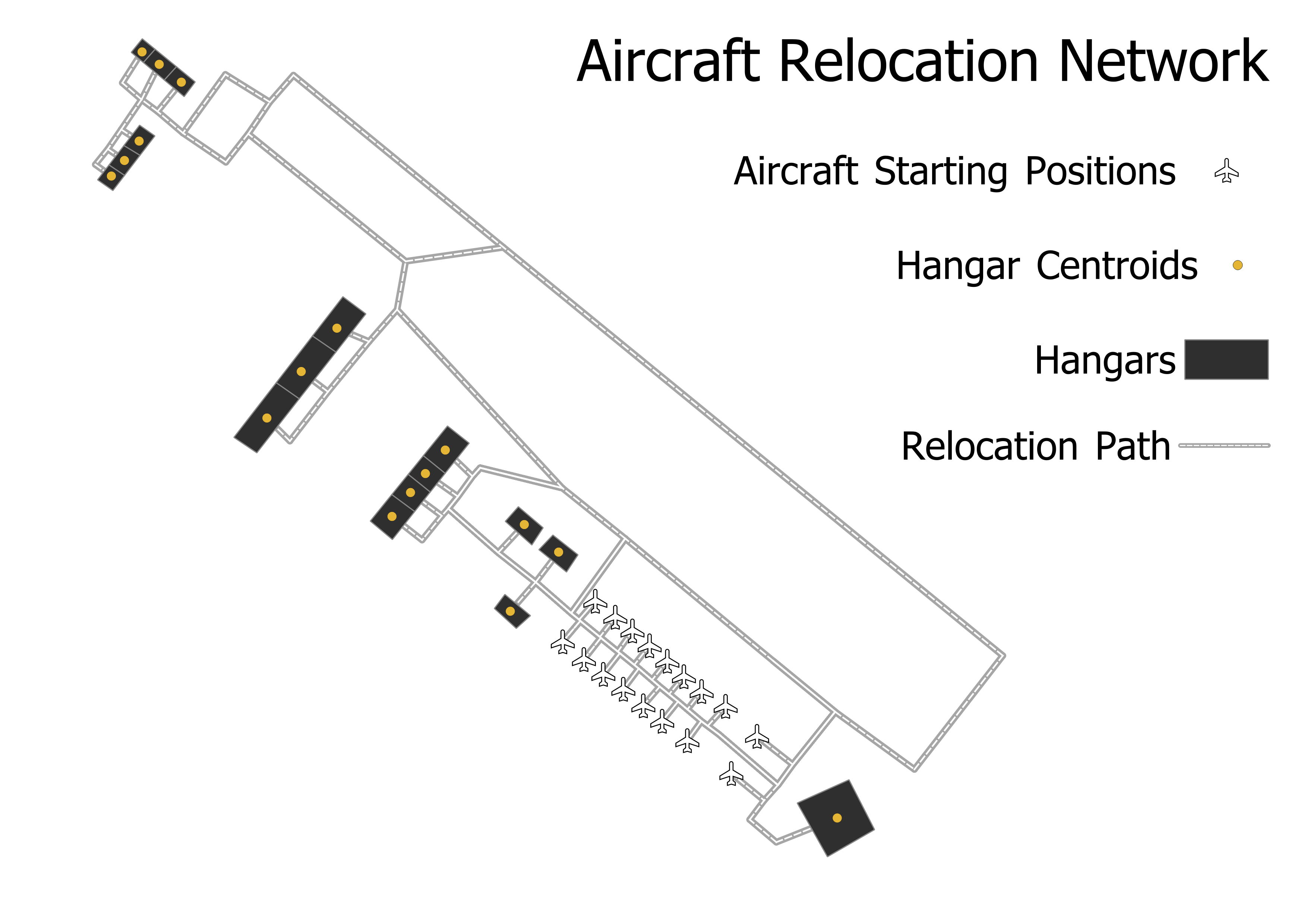}
        \caption{Aircrafts and hangars illustration for the aircraft relocation task.}
        \label{fig:aircraft}
    \end{subfigure}

    \caption{Visual explanations of flood response tasks.}
    \label{fig:tasks}
\end{figure}

In this section we provide details of our decision-making flood response tasks. Specifically, there are four shelters for the evacuation planning task, as shown in Figure \ref{fig:evacuation}. The center of the AFB is selected as the evacuation origin. For the aircraft relocation task, there are 17 aircrafts and 17 hangars under consideration, which are illustrated in Figure \ref{fig:aircraft}. 

\subsection{Implementation Details}
We set $K=4$ to match the four USGS sensor placement in the region of interest. We set batch size to $64$ and time window to $10$, and use Adam optimizer with learning rate $5\times10^{-4}$. We pre-train our model for $50$ epochs for reconstruction only, and then train end-to-end for up to $500$ epochs. For the scoring network, hidden dimension is set to $64$. The spatio-temporal neural network is consisted of a single GCN layer with hidden dimension $16$ followed by a single-layer GRU with hidden dimension $16$ and a linear head. In terms of sensor selection with IMLE, noise is \emph{Sum-of-Gamma} with $k{=}10$, while the target distribution applies $\alpha{=}1.0$ and $\beta{=}10$, and noise temperatures are set to $1.0$ and we draw $1$ sample per instance. During model training, we sample $k\in[1,K]$ per batch, and at validation $K$ remains fixed. In terms of the implementation of the decision network, for the aircraft relocation task, flood impact at hangars averages the $10$ nearest cells with weights $\propto 1/d^{2}$, and we apply 10 sinkhorn iterations for a doubly-stochastic soft assignment. For the evacuation task, we allocate $3200$ people via an MLP with hidden dimension $128$, with overflow penalized above capacity with weight $\lambda=10$.

\subsection{Baseline Methods}
We compare our end-to-end method against a combination of baselines spanning three categories: \emph{sensor selection strategies}, \emph{flood imputation methods}, and \emph{downstream decision-making algorithms}. Below, we briefly summarize the representative baseline methods used in our evaluation.

\noindent\textbf{\emph{- Sensor selection strategies:}}

\textbf{USGS.} This is the baseline sensor selection strategy where sensors are placed in the locations where USGS stream gages were available during the March 2019 flood \cite{usgs_nwis}. 


\textbf{Principal Component Analysis (PCA).} A commonly used unsupervised sensor placement method -- the most informative sensor locations are identified via PCA~\cite{abdi2010principal} based on fitting a d-dimensional PCA with satellite features. 

\noindent\textbf{\emph{- Flood imputation methods:}}

\textbf{Inverse Distance Weighting (IDW).} A classical spatial interpolation method~\cite{lu2008adaptive} that estimates unobserved values based on a weighted average of nearby observations using inverse distance weighting over cell center coordinates.

\textbf{K-Nearest Neighbors (KNN).}
KNN~\cite{peterson2009k} estimates the unobserved values by averaging observations from the $k$ nearest selected sensors in Euclidean space, using distance-based weights. Here for each target cell we query its $k$ closest observed sites and compute inverse-distance weighted average.

\noindent\textbf{\emph{- Downstream decision-making methods:}}

\textbf{Integer Linear Programming (ILP).}
ILP~\cite{graver1975foundations} formulates discrete decision-making as a combinatorial optimization problem under linear constraints. For the evacuation task, we minimize total route cost with one integer variable per route, enforce a demand-balance equality and a per–route capacity, and return the optimal per-route assignments. For the aircraft relocation task, we minimize total assignment cost over binary variables with constraints that each aircraft is assigned to exactly one hangar and each hangar receives at most one aircraft, returning a binary assignment matrix.

\textbf{Inverse-Weighted (IW).}
Given a cost vector over candidate options, IW~\cite{ahuja2001inverse} assigns resources in proportion to inverse costs. For evacuation, we compute inverse-cost weights to allocate total demand, capping each route by capacity. For aircraft relocation, we compute inverse-cost scores and greedily assign each aircraft to the highest-scoring unassigned hangar, yielding a one-to-one assignment.




\subsection{Experimental Results}
We present the quantitative results for both evacuation planning and aircraft relocation tasks in Table~\ref{tab:main_results}. Our proposed method, \textbf{DFL-ST-IMLE}, consistently outperforms all baselines across both decision costs and reconstruction accuracy.
\begin{table}[htbp]
\centering
\small
\setlength{\tabcolsep}{4pt}  
\caption{Comparison with baseline methods on evacuation planning \textit{(top)} and aircraft relocation \textit{(bottom)} tasks. \footnotesize Metrics include total evacuation/relocation costs, average overflow beyond shelter/hangar capacity, prediction MSE, and inference time.}
\label{tab:main_results}
\begin{tabular}{c|c|c|c|c}
\toprule
\textbf{Method} & \makecell[c]{\textbf{Evac.}\\\textbf{Cost ↓}} & \makecell[c]{\textbf{Over}\\\textbf{-flow ↓}} & \makecell[c]{\textbf{Pred.}\\\textbf{MSE ↓}} & \makecell[c]{\textbf{Infer.}\\\textbf{Time (s) ↓}} \\
\midrule
USGS-IDW-ILP & 4476.2 & 0 & 1.2033 & 0.2956 \\
USGS-IDW-IW & 5516.8 & 0 & 1.2033 & 0.2922 \\
USGS-KNN-ILP & 4480.4 & 0 & 1.2096 & 0.3056 \\
USGS-KNN-IW & 5522.0 & 0 & 1.2096 & 0.2968 \\
PCA-IDW-ILP & 5586.7 & 0 & 1.4472 & 0.3039 \\
PCA-IDW-IW & 6885.4 & 0 & 1.4472 & 0.2895 \\
PCA-KNN-ILP & 4227.5 & 0 & 1.4532 & 0.3130 \\
PCA-KNN-IW & 5210.2 & 0 & 1.4532 & 0.3019 \\
\rowcolor{gray!20} Ours (DFL-ST-IMLE) & 3831.5 & 1 & 0.9179 & 0.2888 \\
\bottomrule
\toprule
\textbf{Method} & \makecell[c]{\textbf{Reloc.}\\\textbf{Cost ↓}} & \makecell[c]{\textbf{Over}\\\textbf{-flow ↓}} & \makecell[c]{\textbf{Pred.}\\\textbf{MSE ↓}} & \makecell[c]{\textbf{Infer.}\\\textbf{Time (s) ↓}} \\
\midrule
USGS-IDW-ILP & 71.64 & 0 & 1.1722 & 0.5020\\
USGS-IDW-IW & 71.99 & 0 & 1.1722 & 0.4744\\
USGS-KNN-ILP & 71.64 & 0 & 1.1769 & 0.5065\\
USGS-KNN-IW & 71.99 & 0 & 1.1769 & 0.4893\\
PCA-IDW-ILP & 71.64 & 0 & 1.4525 & 0.4977\\
PCA-IDW-IW & 71.99 & 0 & 1.4525 & 0.4756\\
PCA-KNN-ILP & 71.64 & 0 & 1.4532 & 0.5188\\
PCA-KNN-IW & 71.99 & 0 & 1.4532 & 0.4910\\
\rowcolor{gray!20} Ours (DFL-ST-IMLE) & 68.43 & 0 & 0.8690 & 0.0477\\
\bottomrule
\end{tabular}
\end{table}

On the \textit{evacuation planning} task, our method achieves the lowest evacuation cost of 3831.5, outperforming the best  baseline by a margin of 9.37\%. This cost reduction is achieved despite allowing controlled overflow (around 1 person), demonstrating a deliberate trade-off between path utilization and cost minimization that is learned automatically through end-to-end training on NN-based models. Additionally, our method achieves significantly lower reconstruction error with MSE 0.9179, suggesting that the learned sensor placements provide more informative partial observations to support accurate flood state recovery and forecasting. 

For the \textit{aircraft relocation} task, DFL-ST-IMLE yields a relocation cost of 68.43, improving upon all variants of ILP and IW based baselines, with improvements of 4.48\% and 4.95\%, respectively. Importantly, this improvement is achieved without any overflow violations, and with the best reconstruction quality, i.e., with MSE 0.8690. These results validate the ability of our model to integrate flood-aware relocation cost estimation and structural matching constraints in a unified learning pipeline. Notably, our model also achieves the fastest inference time which is 0.0477s, highlighting its scalability for real-time decision support compared to optimization methods.

Notably, the reported overflow value 1 is the average number of people exceeding the capacity; in other words, it corresponds to roughly one person on average beyond the cap across the evaluated instances. This arises because our model uses a continuous relaxation with a \emph{capacity penalty} rather than a \emph{hard constraint}, allowing a small, cost-efficient violation that improves total evacuation cost. If strict feasibility is required, we can impose hard constraints by (i) solving the downstream integer program on the predicted costs, or (ii) applying a cap-and-redistribute projection that clips each route to its capacity and proportionally reassigns any residual demand to non-saturated routes. In both cases, the evacuation cost will increase relative to the softly constrained solution, reflecting the classic cost–feasibility trade-off. We acknowledge this as a limitation of the soft-constraint setting and view the ILP-projected allocations as a practical, hard-feasible companion to our learned policy.

\begin{figure}[htbp]
    \centering
    \includegraphics[width=\linewidth]{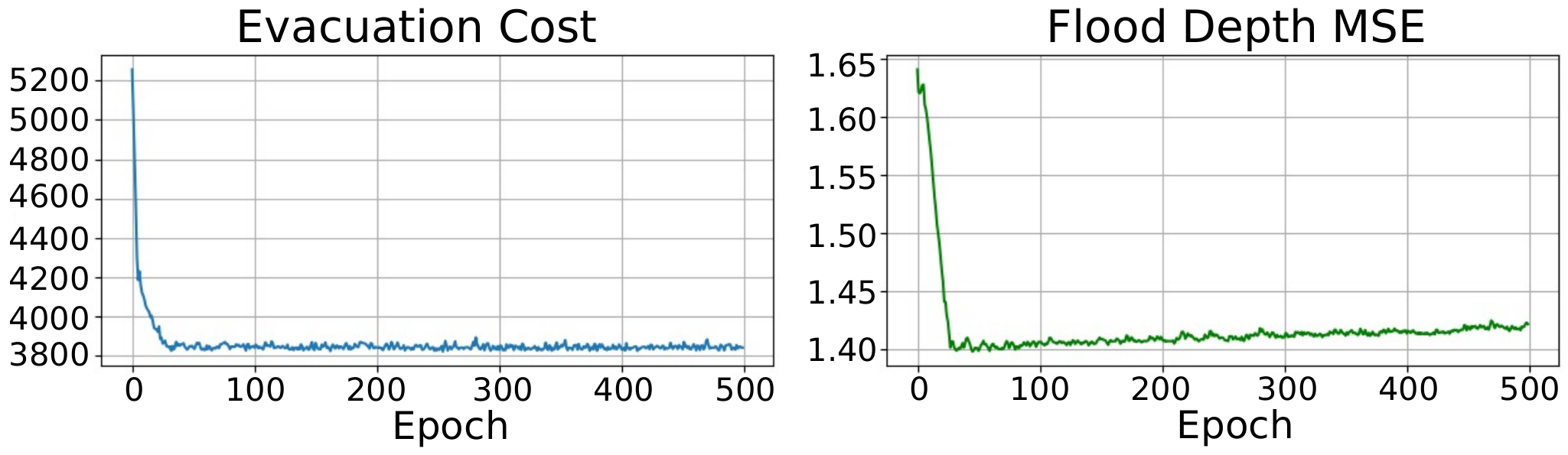}
    \caption{Training curves visualization.}
    \label{fig:training_curve}
\end{figure}

Additionally, we also provide the training curves visualization for the evacuation task in Figure \ref{fig:training_curve}. Overall, the observed gains demonstrate that our framework enables more informative sensor placement, more accurate state estimation, and better downstream decision quality. 

\subsection{Ablation Studies}

To investigate the significance of each module in our framework, we conduct ablation studies by comparing the performance of our model with the following variants:

\noindent\textbf{w/o ST.} This variant replaces our spatio-temporal reconstruction module with a simple linear layer, this variant excludes pre-training with reconstruction loss. 

\noindent\textbf{w/o I-MLE.} This variant replaces our I-MLE top-k sensor selection mechanism with regular top-k sensor selection.

\noindent\textbf{w/o DFL.} This variant is trained end-to-end, replacing the task-specific decision-focused loss with forecasting loss.


\begin{figure}[h]
    \centering
    \includegraphics[width=\linewidth]{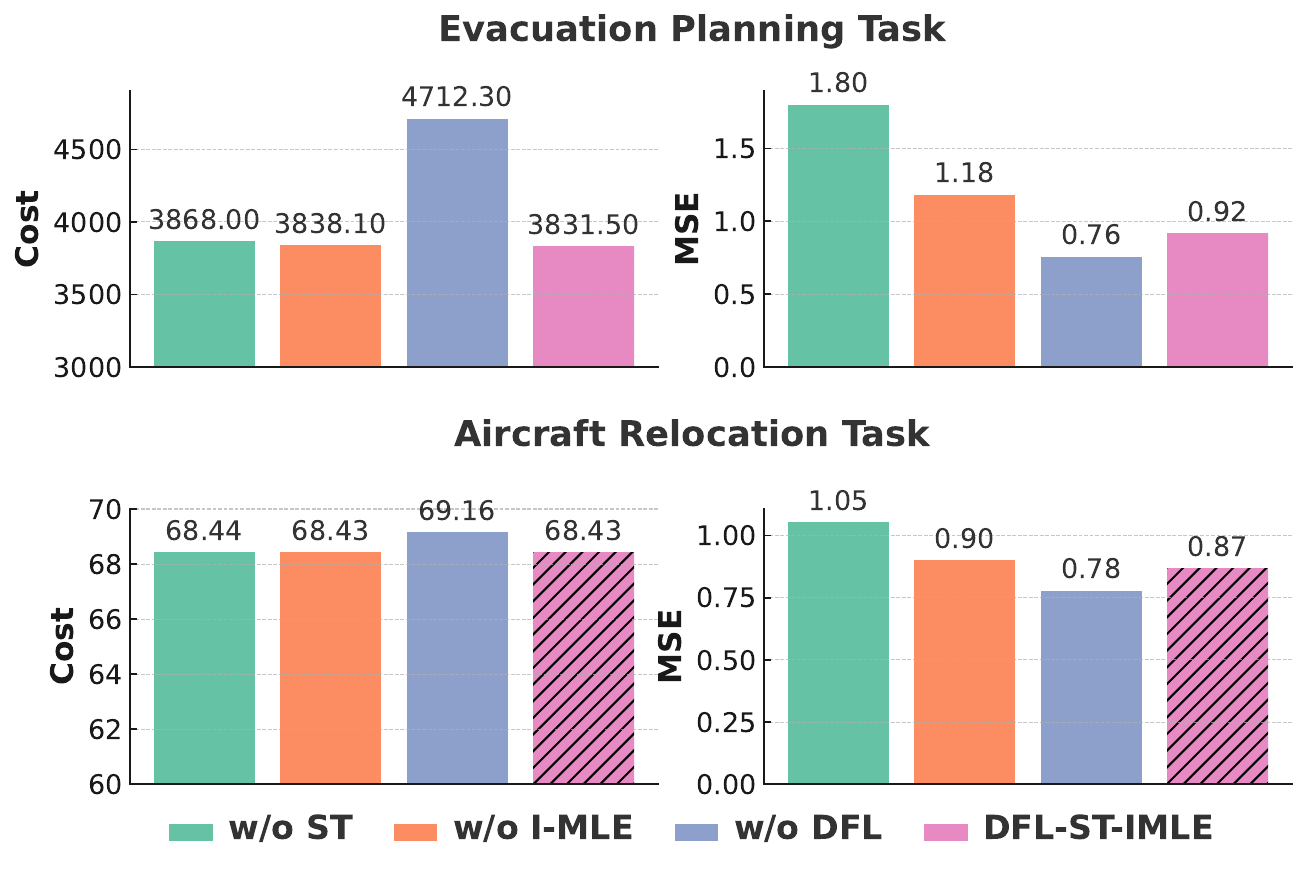}
    \caption{Ablation Studies.}
    \label{fig:ablation}
\vspace{-0.2cm}
\end{figure}

Figure~\ref{fig:ablation} reports the impact of removing key components from our framework. Without the spatio-temporal module (\textbf{w/o ST}), both prediction and decision performance degrade, indicating the importance of structured reconstruction. Removing I-MLE (\textbf{w/o I-MLE}) leads to higher prediction error and slightly worse downstream cost, highlighting the benefit of gradient-informed sensor selection. The most significant drop occurs when removing the decision-focused loss (\textbf{w/o DFL}), which achieves low prediction error but yields substantially higher decision cost. This confirms that minimizing forecasting error alone is insufficient, and task-aligned learning is critical for effective decision-making.
\subsection{Limitations and Future Research}

Our current sensor deployment framework selects optimal sensor locations that are agnostic to where existing USGS stream gauges are installed. This may inflate evaluation performance for baseline methods, since the accuracy of the numerical solver which produces ground-truth hydraulic depth measurements is dependent on the boundary condition locations of the existing sensors that initialized the HEC-RAS model through stage and discharge observations. Through moderate adjustment, our framework could be easily adopted to scenarios where there are pre-existing sensor network and additional sensor location recommendation is needed. Future research focused on robust optimization should evaluate sensor placements across multiple plausible futures and test sites via ensemble forecasting and diverse geographic environments. 
\section{Conclusion}
In this paper, we present an end-to-end decision-focused framework for flood response that jointly learns sensor placement and forecasting with the ultimate goal of improving downstream decisions. By embedding a differentiable decision layer and leveraging I-MLE for sensor selection, our approach aligns sensing and prediction modules directly with task-level performance. Experiments on real-world flood scenarios confirm that our method not only enhances flood state reconstruction but also leads to significantly better evacuation and resource allocation outcomes, highlighting the importance of decision-aware learning in sensor-driven disaster management.
\section{Acknowledgment}
This work is partially supported by NSF CMMI-2242590 and CAREER 2442712. Any mention of commercial products is for informational purposes and does not constitute an endorsement by the US government.

\bibliographystyle{plain}
\bibliography{reference}

\end{document}